# Multiple View Reconstruction of Calibrated Images using Singular Value Decomposition

Ayan Chaudhury, Abhishek Gupta, Sumita Manna, Subhadeep Mukherjee, Amlan Chakrabarti

*Abstract*— **Calibration in a multi camera network has widely been studied for over several years starting from the earlier days of photogrammetry. Many authors have presented several calibration algorithms with their relative advantages and disadvantages. In a stereovision system, multiple view reconstruction is a challenging task. However, the total computational procedure in detail has not been presented before. Here in this work, we are dealing with the problem that, when a world coordinate point is fixed in space, image coordinates of that 3D point vary for different camera positions and orientations. In computer vision aspect, this situation is undesirable. That is, the system has to be designed in such a way that image coordinate of the world coordinate point will be fixed irrespective of the position & orientation of the cameras. We have done it in an elegant fashion. Firstly, camera parameters are calculated in its local coordinate system. Then, we use global coordinate data to transfer all local coordinate data of stereo cameras into same global coordinate system, so that we can register everything into this global coordinate system. After all the transformations, when the image coordinate of the world coordinate point is calculated, it gives same coordinate value for all camera positions & orientations. That is, the whole system is calibrated.**

## I. Introduction

Several camera calibration mechanisms have presented by various authors with their relative advantages and disadvantages. However, all these algorithms concentrate on extracting camera parameters. In the model proposed by Aziz & Karara[1], parameters are extracted in an elegant fashion, though, they neglected lens distortion of the cameras. Later on, a model proposed by Tsai[2] opened a new door to form the basis of accurate camera calibration. But Tsai's algorithm also suffers from non-linear search, depends on the data precision. Homography based methods firstly proposed by Zhang[3] is a modern method and makes use of advanced projective geometry, but here also lots of computational effort is need at initial stage. Besides these traditional methods, a variety of algorithms have been presented in several literatures.

Ayan Chaudhury & Abhishek Gupta are with the department of Computer Science & Engineering, University of Calcutta, 92 A.P.C. Road Kolkata-700009, India. (Phone: +91 9830783652 and +91 9143008407 ; email: ayanchaudhury.cs@gmail.com and abhishek_gupta118@yahoo.com)

Sumita Manna and Subhadeep Mukherjee are with the A.K.Choudhury School of Information Technology, University of Calcutta, 92 A.P.C. Road Kolkata-700009, India. (Phone: +91 9230272856 and +91 8013526896 ; email: sumimca15@gmail.com and soumo12@gmail.com)

Amlan Chakrabarti is with the A.K.Choudhury School of Information Technology, University of Calcutta, 92 A.P.C. Road Kolkata, India (phone: 091-33-23500289/+91 9831129520; fax: 091-33-23519755 ; email: acakcs@caluniv.ac.in)

But in this paper, our aim is not to demonstrate a new calibration method. We are concentrating on the problem of multiple view reconstruction in a multi camera network, where we will make extensive use of the traditional models described so far for the calibration of cameras.

Let, the same static scene be imaged by two cameras C & C' as shown in figure1.

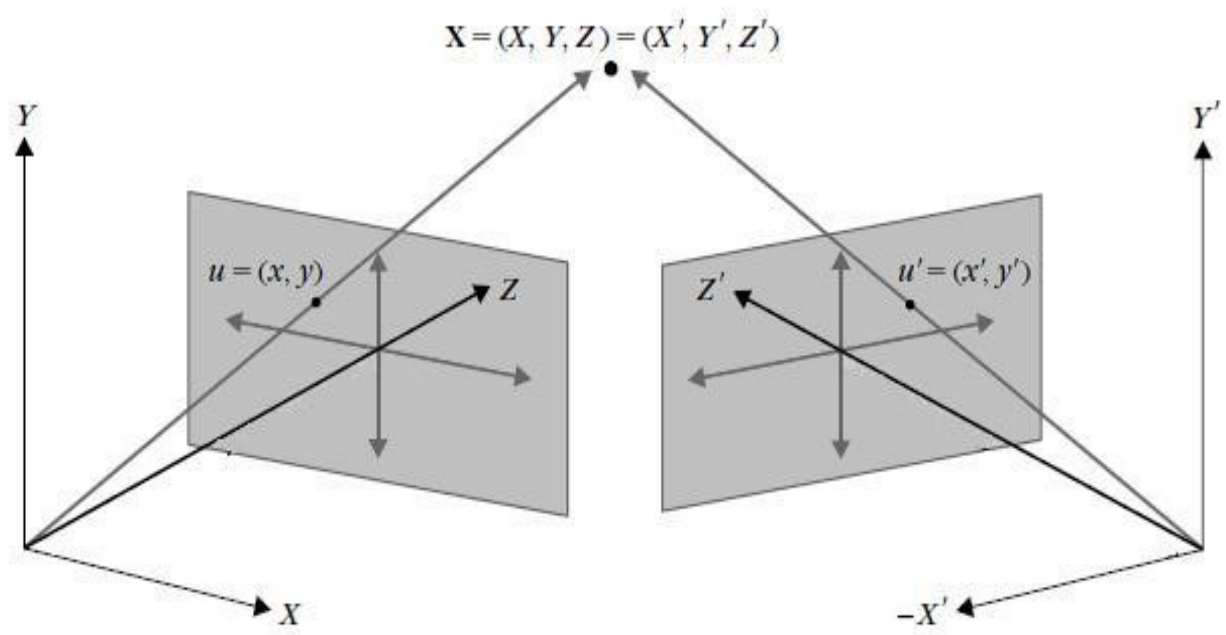

Fig.1 Imaging of a static scene by two cameras

These could be two physically separate cameras or a single moving camera at different positions. Let the scene coordinates of a point X in the C coordinate system be (X,Y,Z) and in the C' coordinate system be (X',Y',Z'). We denote corresponding image coordinates of X in image plane P and P' by u = (x,y) and u' = (x',y'). The points u and u' are said to be corresponding points.

Hence, the same world coordinate scene is mapped at different image coordinates. Our aim is to relate corresponding points mathematically by an explicit one-to-one mapping. That is, after calibration of the cameras, for same world coordinate scene, every camera will image to the same pixel. And this should be true for an arbitrary number of cameras in the network.

For simplicity matters, we restrict to a single calibration object in the static scene and would verify our result on that point.

## II. Proposed Model

Figure2 shows the proposed calibration method. In our method, we have concentrated on our ultimate goal to 'unify' different camera views. To keep matters simple we have neglected lens distortions, though it can be incorporated using the technique proposed by Shih[4].

In the first phase of our work, we have generated a calibration pattern having 19 calibration points, whose coordinates in local and global coordinate system are being measured.

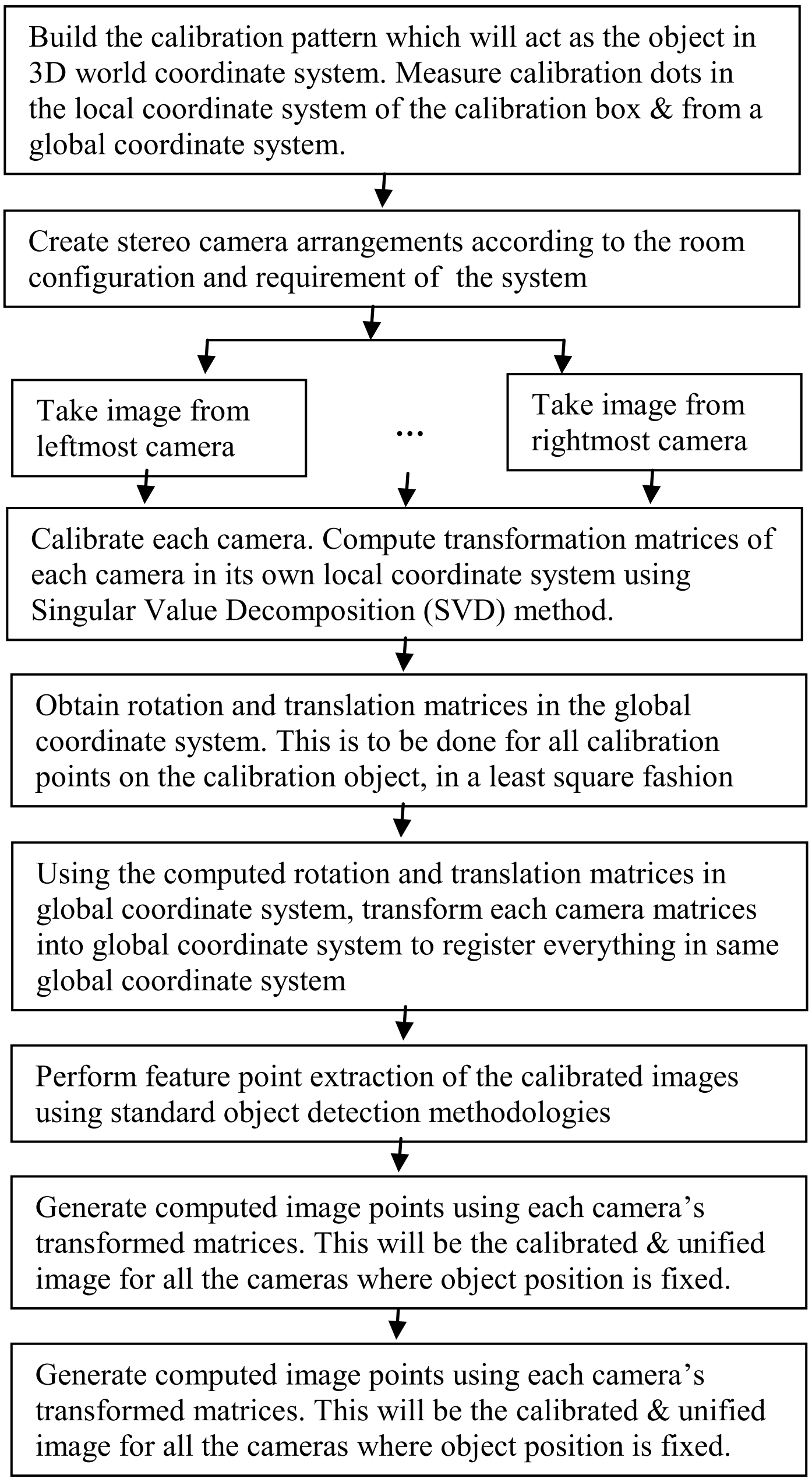

Fig. 2 A scheme of the proposed calibration method

Out of these dots, eight are used in extracting camera parameters and 18 are used for computing global coordinate data. Firstly, from each camera position snapshots are taken and pixel coordinates of predetermined six points are measured. Then from 3D-2D coordinates, camera parameters of all cameras in their own local coordinate system are computed. Then we use global coordinate data to transfer all local coordinate data to same global coordinate to register everything into same global coordinate system.

We then perform some morphological image processing operations to extract feature points from each image taken by the stereo cameras. This step is necessary for further processing with the images. After all these operations are done, image coordinates as well as pixel coordinates of the fixed calibration point in space is computed. It is seen that although image coordinates of a fixed world coordinate point varies for different camera positions and orientations, we have calibrated the system in such a way that image coordinate of this point does not vary with camera positions and orientations. That is, the whole system is being calibrated.

## III. Mathematical Formulation

Now we present the whole mathematical formulation of the system. The first phase consists of extraction of camera parameters and next phase is coordinate system transformation.

### *A. Extraction of camera parameters*

In order to express an arbitrary object point in the world coordinate system, we first need to transform it to camera coordinates. This transformation consists of a translation and a rotation. Let, a world coordinate point be represented as $P_w(x_w,y_w,z_w)$ and corresponding camera 3D coordinate be $C(x,y,z)$. Then the transformation from 3D world to 3D camera coordinate can be represented as:

$$\begin{bmatrix} x \\ y \\ z \end{bmatrix} = \begin{pmatrix} R_{11} & R_{12} & R_{13} \\ R_{21} & R_{22} & R_{23} \\ R_{31} & R_{32} & R_{33} \end{pmatrix} \begin{bmatrix} X_w \\ Y_w \\ Z_w \end{bmatrix} + \begin{pmatrix} T_x \\ T_y \\ T_z \end{pmatrix} = \begin{pmatrix} R_{11}X_w + R_{12}Y_w + R_{13}Z_w + T_x \\ R_{21}X_w + R_{22}Y_w + R_{23}Z_w + T_y \\ R_{31}X_w + R_{32}Y_w + R_{33}Z_w + T_z \end{pmatrix} = \begin{bmatrix} R_1^T P_w + T_x \\ R_2^T P_w + T_y \\ R_3^T P_w + T_z \end{bmatrix} \quad (1)$$

Let us denote the following:

$$X = R_{11}X_w + R_{12}Y_w + R_{13}Z_w + T_x$$
$$Y = R_{21}X_w + R_{22}Y_w + R_{23}Z_w + T_y$$
$$Z = R_{31}X_w + R_{32}Y_w + R_{33}Z_w + T_z$$

If f be the focal length of the camera, then from pinhole geometry we can write,

$$x = f\frac{X}{Z} \quad \& \quad y = f\frac{Y}{Z} \quad (2)$$

The intrinsic camera parameters usually include the effective focal length f, scale factor $s_u$, and the image center $(u_0, v_0)$ also called the principal point. Here, as usual in computer vision literature, the origin of the image coordinate system is in the upper left corner of the image array. The unit of the image coordinates is pixels.

Let, $(x_{im}, y_{im})$ be the pixel coordinates & $(O_x, O_y)$ be the optical center. If we consider scaling along x and y directions are considered as $S_x$ and $S_y$ then the following equation can be written:

$$x = (x_{im} - O_x)\, S_x \quad \& \quad y = (y_{im} - O_y)\, S_y \quad (3)$$

Hence, from last two equations we can write,

$$\frac{f}{S_x}\frac{X}{Z} = x_{im} - O_x \quad \& \quad \frac{f}{S_y}\frac{Y}{Z} = y_{im} - O_y \quad (4)$$

which implies, $x_{im} = \alpha_u \frac{X}{Z} + O_x \quad \& \quad y_{im} = \alpha_v \frac{Y}{Z} + O_y \quad (5)$

where $\alpha_u$ and $\alpha_v$ are considered as parameters for scaling in x and y directions respectively.

The pinhole model is only an approximation of the real camera projection. It is a useful model that enables simple mathematical formulation for the relationship between object and image coordinates. However, it is not valid when high accuracy is required and therefore, a more comprehensive camera model must be used. Usually, the pinhole model is a basis that is extended with some corrections for the systematically distorted image coordinates. To keep things simple, we neglect lens distortion.

So, in the first stage of the mathematical formulation, we have done the transformation : from world coordinate to pixel coordinate. Now, our aim is to compute the parameters. So, we will use Direct Linear Transformation(DLT) method[1]. The DLT method is based on the pinhole camera model and it ignores the nonlinear radial and tangential distortion components.

In this method, a linear transformation equation can be written to map world coordinate $(x_w, y_w, z_w)$ to pixel coordinate (x,y) as:

$$\begin{bmatrix} x \\ y \\ 1 \end{bmatrix} \sim M \begin{bmatrix} x_w \\ y_w \\ z_w \\ 1 \end{bmatrix} \quad \text{which can be written as,}$$

$$\begin{bmatrix} \alpha x \\ \alpha y \\ \alpha \end{bmatrix} = M \begin{bmatrix} x_w \\ y_w \\ z_w \\ 1 \end{bmatrix} \qquad (6)$$

where, α is a parameter and M is the calibration matrix given by

$$M = \begin{pmatrix} m_{11} & m_{12} & m_{13} & m_{14} \\ m_{21} & m_{22} & m_{23} & m_{24} \\ m_{31} & m_{32} & m_{33} & m_{34} \end{pmatrix} \qquad (7)$$

which consists of camera internal and external parameters. Hence, equation (6) can be rewritten as,

$$\begin{bmatrix} x \\ y \\ 1 \end{bmatrix} \sim M_{int} M_{ext} \begin{bmatrix} x_w \\ y_w \\ z_w \\ 1 \end{bmatrix} \qquad (8)$$

Where $M_{int}$ represents the matrix containing internal parameters and $M_{ext}$ represents the matrix containing external parameters.

*1) Solving Through Least Square Approach:* Upto this point, we have only estimated the equation for solving intrinsic & extrinsic camera parameters. Now we have to solve equation for obtaining M, the calibration matrix.

Let's refer to the equation (6). We can obtain value of α as

$$\alpha = m_{31}x_w + m_{32}y_w + m_{33}z_w + m_{34}$$

Substituting this value back in the equation & rearranging the equations, we obtain

$$m_{11}x_w + m_{12}y_w + m_{13}z_w + m_{14} - x\, m_{31}x_w - x\, m_{32}y_w - x\, m_{33}z_w - x\, m_{34} = 0 \qquad (9)$$

$$m_{21}x_w + m_{22}y_w + m_{23}z_w + m_{24} - y\, m_{31}x_w - y\, m_{32}y_w - y\, m_{33}z_w - y\, m_{34} = 0 \qquad (10)$$

So, we have two equations and 12 unknowns $(m_{11},\ldots,m_{34})$. These equations can be solved in an elegant fashion. We can make the system as a overdetermined system of linear equations and use the property of statistical model fitting. Here we have taken 8 calibration points, whose 3D-2D coordinates are known. However, many more correspondences & equations can be obtained and M can be estimated through least square technique. If we assume we are given N matches for the homogeneous linear system, we have the following equation:

$$\mathbf{L}\mathbf{a} = 0 \qquad (11)$$

where

L=

$$\begin{bmatrix}
X_1 & Y_1 & Z_1 & 1 & 0 & 0 & 0 & 0 & -X_1u_1 & -Y_1u_1 & -Z_1u_1 & -u_1 \\
0 & 0 & 0 & 0 & X_1 & Y_1 & Z_1 & 1 & -X_1v_1 & -Y_1v_1 & -Z_1v_1 & -v_1 \\
\vdots & \vdots & \vdots & \vdots & \vdots & \vdots & \vdots & \vdots & \vdots & \vdots & \vdots & \vdots \\
X_i & Y_i & Z_i & 1 & 0 & 0 & 0 & 0 & -X_iu_i & -Y_iu_i & -Z_iu_i & -u_i \\
0 & 0 & 0 & 0 & X_i & Y_i & Z_i & 1 & -X_iv_i & -Y_iv_i & -Z_iv_i & -v_i \\
\vdots & \vdots & \vdots & \vdots & \vdots & \vdots & \vdots & \vdots & \vdots & \vdots & \vdots & \vdots \\
X_N & Y_N & Z_N & 1 & 0 & 0 & 0 & 0 & -X_Nu_N & -Y_Nu_N & -Z_Nu_N & -u_N \\
0 & 0 & 0 & 0 & X_N & Y_N & Z_N & 1 & -X_Nv_N & -Y_Nv_N & -Z_Nv_N & -v_N
\end{bmatrix}$$

and $\mathbf{a} = [\, m_{11}, m_{12}, \ldots, m_{33}, m_{34}\,]^T$

We can estimate the parameters in a least square fashion. In order to avoid a trivial solution $m_{11},\ldots, m_{34} = 0$, a proper normalization must be applied. Abdel-Aziz and Karara[1] used the constraint $m_{34} = 1$. Then, the equation can be solved with a pseudoinverse technique. The problem with this normalization is that a singularity is introduced, if the correct value of $m_{34}$ is close to zero. Faugeras & Toscani[5] suggested the constraint $a_{31}^2 + a_{32}^2 + a_{33}^2 = 1$ which is singularity free.

*2) Extraction of Parameters through Singular Value Decomposition (SVD):* So far we have structured the equation containing DLT matrix and structured it through least square technique. But, the main problem seem to be unsolved till now : extraction of parameters from M.

In an overdetermined system, solution can be get through SVD. In this method, a singular matrix can be partitioned as:

$$A = U\, W\, V^T$$

& because vector V gives the solution corresponding to smallest eigenvalue, it gives the real solution. Similar case we apply for matrix L. That is, by singular value decomposition of matrix L, we obtain matrix M. Then we will extract parameters from M.

It can be seen from (8), M consists of the combination of internal & external calibration matrices. There are techniques for extracting some of the physical camera parameters from the DLT matrix, but not many are able to solve all of them. In order to extract parameters from DLT matrix, we prefer the system proposed by Melen[6] where he had done QR decomposition to obtain the following :

$$M = \lambda\, V^{-1} B^{-1} F\, R\, T \qquad (12)$$

Where λ is a overall scaling factor and R , T are the rotation and translation matrices from the object coordinate system to the camera coordinate system. Matrices V , B and F contain the focal length f, principal point $(u_0, v_0)$ and coefficients for the linear distortion $(b_1, b_2)$:

$$V=\begin{bmatrix} 1 & 0 & -u_0 \\ 0 & 1 & -v_0 \\ 0 & 0 & 1 \end{bmatrix};\ B=\begin{bmatrix} 1+b_1 & b_2 & 0 \\ b_2 & 1-b_1 & 0 \\ 0 & 0 & 1 \end{bmatrix};\ F=\begin{bmatrix} f & 0 & 0 \\ 0 & f & 0 \\ 0 & 0 & 1 \end{bmatrix}$$

Combining expression for F with λ (refer to equation 5) we can write:

$$F=\begin{bmatrix}\alpha_u & 0 & 0\\ 0 & \alpha_v & 0\\ 0 & 0 & 1\end{bmatrix} \text{ so that } V^{-1}F = \begin{bmatrix}\alpha_u & 0 & u_0\\ 0 & \alpha_v & v_0\\ 0 & 0 & 1\end{bmatrix} \quad (13)$$

Hence, equation (13) represents the matrix representing camera internal parameters as far as we are neglecting lens distortion coefficient. Hence, we can write the following:

$$M_{ext} = \begin{bmatrix} r_{11} & r_{12} & r_{13} & T_x\\ r_{21} & r_{22} & r_{23} & T_y\\ r_{31} & r_{32} & r_{33} & T_z\end{bmatrix} \ \& \ M_{int} = \begin{bmatrix}\alpha_u & 0 & u_0\\ 0 & \alpha_v & v_0\\ 0 & 0 & 1\end{bmatrix}$$

Multiplying & from (8) we write expression for M as

$$M = \begin{bmatrix} \alpha_u r_{11} + u_0 r_{31} & \alpha_u r_{12} + u_0 r_{32} & \alpha_u r_{13} + u_0 r_{33} & \alpha_u T_x + u_0 T_z\\ \alpha_v r_{21} + v_0 r_{31} & \alpha_v r_{22} + v_0 r_{32} & \alpha_v r_{23} + v_0 r_{33} & \alpha_v T_y + v_0 T_z\\ r_{31} & r_{32} & r_{33} & T_z \end{bmatrix}$$

$$= \begin{bmatrix} m_1^T & m_{14}\\ m_2^T & m_{24}\\ m_3^T & m_{34}\end{bmatrix} \quad (14)$$

After this stage, we will impose constraint norm(m3) = ‖ m3‖ = 1 and $m_{34}$ = 1 & write equation for M as vector form as:

$$M = \begin{bmatrix} \alpha_u r_1 + u_0 r_3 & \alpha_u T_x + u_0 T_z\\ \alpha_v r_2 + v_0 r_3 & \alpha_v T_y + v_0 T_z\\ r_3 & T_z\end{bmatrix} = \begin{bmatrix} m_1^T & m_{14}\\ m_2^T & m_{24}\\ m_3^T & m_{34}\end{bmatrix} \quad (15)$$

From the above equation, clearly we have,

$$T_z = m_{34} \quad \text{and} \quad r_3 = m_3^T \quad (16)$$

Now we compute following dot and cross products to get a solution:

$$m_1^T \cdot m_3 = (\alpha_u r_1 + u_0 r_3) \cdot r_3 = \alpha_u r_1 \cdot r_3 + u_0 r_3 \cdot r_3 = u_0 \quad (17)$$

Similarly, $m_2^T \cdot m_3 = v_0$ (18)

Now, to compute $\alpha_u$ and $\alpha_v$ we do the following :

$m_1^T \cdot m_1 = (\alpha_u r_1 + u_0 r_3) \cdot (\alpha_u r_1 + u_0 r_3) = \alpha_u^2 + u_0^2$ so that

$$\alpha_u = \sqrt{m_1^T m_1 - u_0^2} \ \ \& \ \ \alpha_v = \sqrt{m_2^T m_2 - v_0^2} \quad (19)$$

Again, from equation (14), we can write the following:

$$\alpha_u r_1 + u_0 r_3 = m_1^T \ \Rightarrow \ \alpha_u r_1 = m_1^T - u_0 m_3^T$$

giving $r_1 = (m_1^T - u_0 m_3^T)/\alpha_u$ (20)

and $r_2 = (m_2^T - v_0 m_3^T)/\alpha_v$ (21)

Again, as $\alpha_u T_x + u_0 T_z = m_{14}$ we obtain

$T_x = (m_{14} - u_0 m_{34})/\alpha_u$ (22)

and $T_y = (m_{24} - v_0 m_{34})/\alpha_v$ (23)

Now, R obtained in this way may not be orthogonal so that we compute another decomposition & get $R = UDV^T$ & do $R = U * V^T$

*B. Coordinate System Transformation*

So far we have calibrated each camera and extracted parameters in their own local coordinate system. Now, in order to achieve our goal to calibrate a point in space irrespective of the camera positions, we follow the approach proposed by Xiong & Queck[7]. In this approach, after calibrating all cameras, we need transfer all of these local coordinate systems into a global coordinate system, so that we can register everything into the same coordinate system. During the calibration process, besides taking calibration points for extracting parameters, we have measured coordinates of 18 points in local coordinate system and global coordinate system. The aim is to mathematically calculate rotation and translation matrices which can transform all calibration points to global coordinate system. Then we will apply these matrices to all camera matrices to register everything in same global coordinate system.

Let Y denote the global coordinate system X denote the local one. Then transformation from X to Y can be done through the following equation:

$$Y = R X + T \quad (24)$$

where R is the rotation matrix and T is the translation matrix.

Suppose we have n global calibration points & their positions are:

$$X = (X_1 \ X_2 \ldots \ X_n)^T$$

where $X_1 = (x_{11} \ x_{12} \ x_{13})$, ... , $X_n = (x_{n1} \ x_{n2} \ x_{n3})$

Similarly for Y coordinate we also have,

$$Y = (Y_1 \ Y_2 \ldots \ Y_n)^T$$

where $Y_1 = (y_{11} \ y_{12} \ y_{13})$, ... , $Y_n = (y_{n1} \ y_{n2} \ y_{n3})$

From above equations we have,

$$\begin{pmatrix} y_{21}-y_{11} & y_{22}-y_{12} & y_{23}-y_{13}\\ y_{31}-y_{11} & y_{32}-y_{12} & y_{33}-y_{13}\\ \vdots & \vdots & \vdots\\ . & . & .\\ y_{n1}-y_{11} & y_{n2}-y_{12} & y_{n3}-y_{13}\end{pmatrix} = \begin{pmatrix} x_{21}-x_{11} & x_{22}-x_{12} & x_{23}-x_{13}\\ x_{31}-x_{11} & x_{32}-x_{12} & x_{33}-x_{13}\\ \vdots & \vdots & \vdots\\ . & . & .\\ x_{n1}-x_{11} & x_{n2}-x_{12} & x_{n3}-x_{13}\end{pmatrix} \begin{pmatrix} r_{11} & r_{21} & r_{31}\\ r_{12} & r_{22} & r_{32}\\ r_{13} & r_{23} & r_{33}\end{pmatrix} \quad (25)$$

Above equation can be expressed as:

$$AZ = b \quad (26)$$

This is an over-determined problem. We use least square approach to solve it,i.e.,

$$e = \sum_{k=1}^{3}\sum_{i=1}^{n-1}\left(b_{ik} - \sum_{j=1}^{3} a_{ij} Z_{jk}\right)^2 = \| b - A Z \|_2^2 \quad (27)$$

By minimizing eqation (27) we get R and with equation (24) and $X_1$ , $Y_1$ we can obtain T as

$$T = Y_1 - R X_1 \quad (28)$$

Once we get R and T , we transform local coordinate system to global coordinate system.

## IV. Experimental Result

After all the theoretical work-ups, we have done the experimental verification. We have made a small experimental setup, where a white sheet of paper glued into a board having black dots painted on it acted as the object. Four snapshots are taken from different positions.The images are shown in Fig2. The coordinates of calibration points are measured in local & global coordinate system. Corresponding 2D pixel coordinates are also measured. Then the total computation is simulated in MATLAB and camera parameters in each camera's local coordinate system are obtained. We have not considered lens distortion into consideration. But this has not hampered our result largely. After back projection of matrix M of each cameras to 2D pixel coordinates, the mean error calculated is shown in TableI.

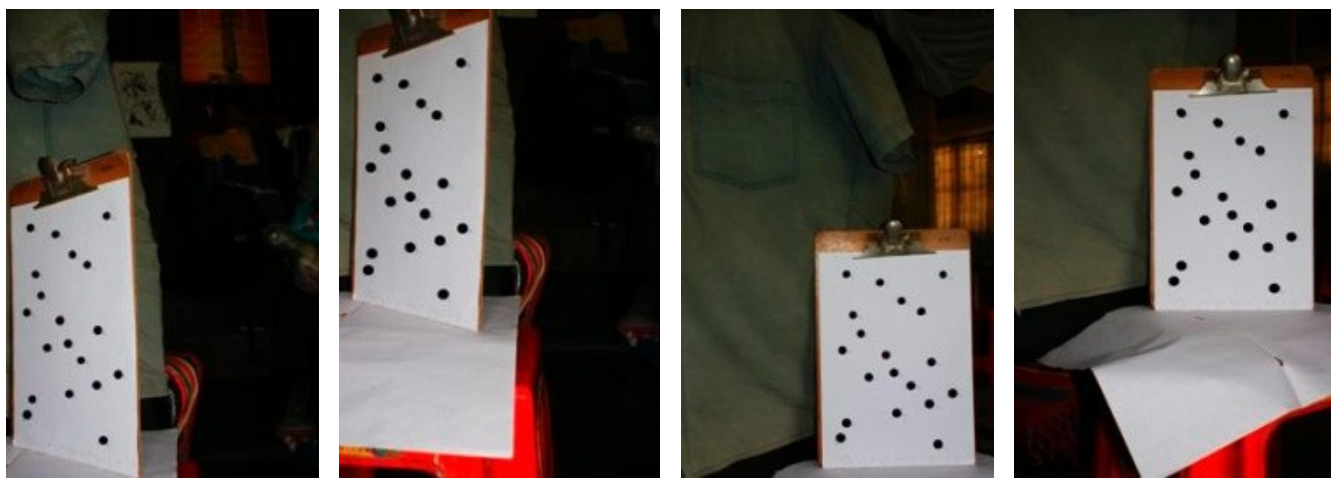

(a) (b) (c) (d)

Fig.2 Snapshot of calibration object from four camera positions

TABLE I
MEAN ERRORS FOR EACH CAMERA POSITION

| **Camera positions** | **Mean error in x-direction** | **Mean error in y-direction** |
|---|---|---|
| 1 | -0.0259 | -0.1195 |
| 2 | -0.0861 | -0.0741 |
| 3 | 0.0213 | -0.0378 |
| 4 | -0.8733 | -0.8122 |

Through least square approach we have computed rotation & translation matrices in global coordinate system and using these values transformed R and T matrices of each camera(which are in their own local coordinate system) to same global coordinate system.

So, we have registered everything into same reference frame.

*1) Feature Point Extraction:* So far we have performed mathematical transformations to compute desired result. Now we perform one of the basic tasks in a stereovision system: feature point extraction through object detection. We have done it through some morphological image processing operations. The processed images are shown in Fig(3a-e).

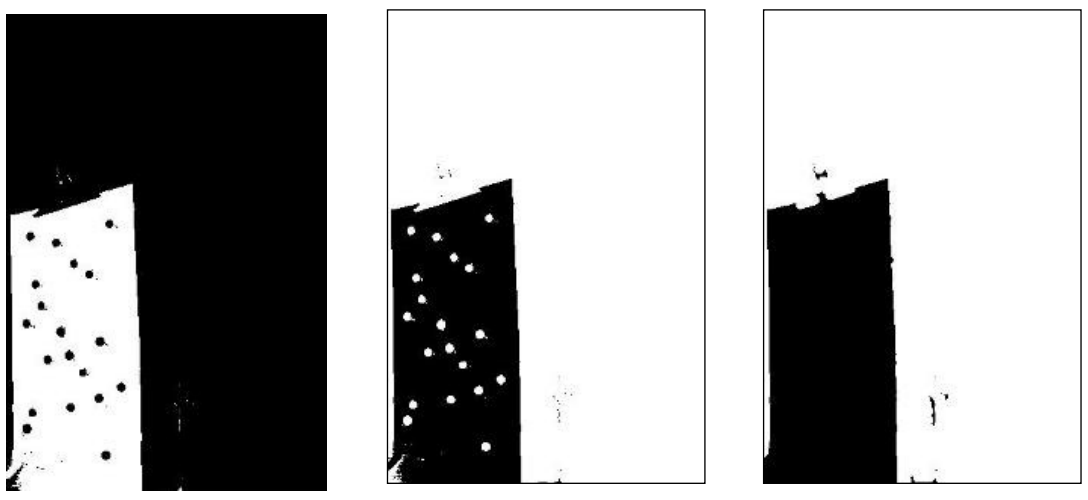

(a) (b) (c)

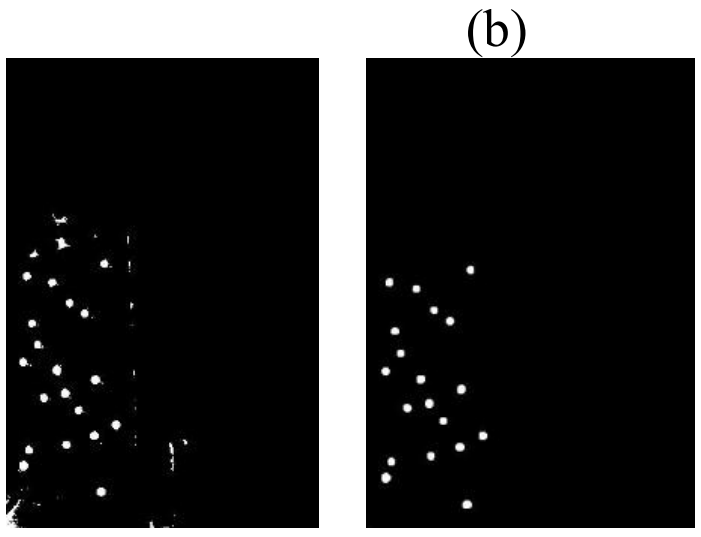

(d) (e)

Fig.3 Processed images in performing object detection

Here we are showing processing of one image. In the sequence of images, the first one is the binary version of the colour image. Next, we've inverted it(image (b) in the sequence). After then, the background is extracted in third image and in fourth of the sequence, background is being subtracted from inverted one. At this stage we've got detected object with some redundancies (unexpected object-like features). So we've performed another step to detect the object from this image(determining pixel connectivity). The last image in the sequence is the final one where white calibration points are detected. We have done same operation for other images also. The finally processed images are shown in Fig4(a-d).

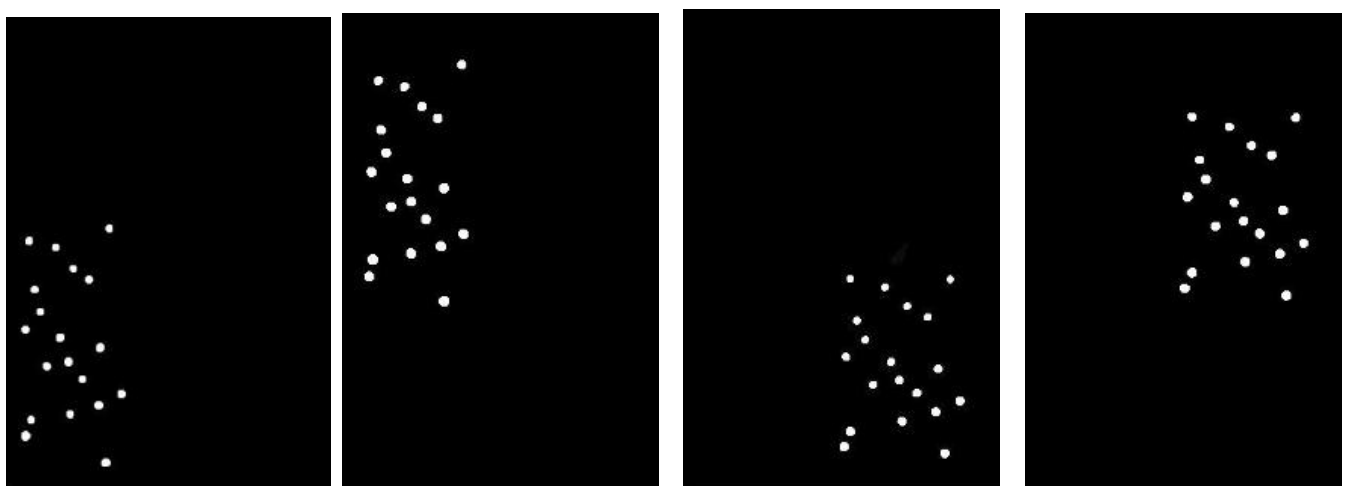

(a) (b) (c) (d)

Fig.4 Sequences of images after object detection , taken from four camera positions

*2) Verification of the results:* Finally we are left with getting the calibrated image coordinate, where the verification will be done, i.e., getting same image coordinate values for different camera positions.

After the transformation from all camera's parameters from its local coordinate to same global coordinate, we have computed image coordinate from each camera's data and with no surprise, after rounding off we have got same image coordinate values for all camera! The coordinate value got in the computation is (158,88).

Hence, for a single calibration point, we have successfully calibrated the system. The same simulation can be done for all points in the image. For simplicity we have considered a single point.

In the next step we have converted image coordinate to pixel coordinate by the following formula as from[2]:

$$X_f = S_x \, d_x'^{-1} X_d + C_x$$
$$Y_f = d_y^{-1} Y_d + C_y$$

where $C_x$ and $C_y$ are the row and column numbers of the centre of computer frame memory , $d_x$ is the centre to centre distance between adjacent sensor elements in X (scan line) direction , $d_y$ is centre to centre distance between adjacent CCD sensor in the Y direction , $S_x$ is the uncertainty factor.

Here we have made some sort of simplifications. We have neglected the effect of uncertainty factor $S_x$ . According to Tsai[2], this factor is introduced due to a variety of factors such as slight hardware timing mismatch between image acquisition hardware and camera scanning hardware etc. Because we are doing the experiment with a single camera, we can consider that the effect will be same for all images, yielding same pixel value.

The image center in the pinhole camera model is the point in the image plane at the base of the line that is perpendicular to the image plane, passing through the focal point. Some interesting facts about image center approximation can be found from Tapper[8]. Tsai initially recommended using center of the frame buffer as a reasonable estimate. After further experimentation, he found that the image center in modern CCD cameras often varies so widely from this estimate that accurate calibration is impossible.
According to Tapper[8], image center can be investigated as finding the orthocenter of the vanishing points of three orthogonal sets of parallel lines. However, this process is very cumbersome and is not accurate enough. So, we have decided to take the image center as the mid point of the pixel array, that is, for a M × N image, the image center is taken as the point $(\frac{M}{2}, \frac{N}{2})$. Here we have taken a (1000 x 1100) image, so that we have taken image center as the point(500,550).
After estimating everything, we have implemented the calibration object to pixel coordinate and reconstructed the other points through relative pixel shift. The final image is Shown in Fig(5).

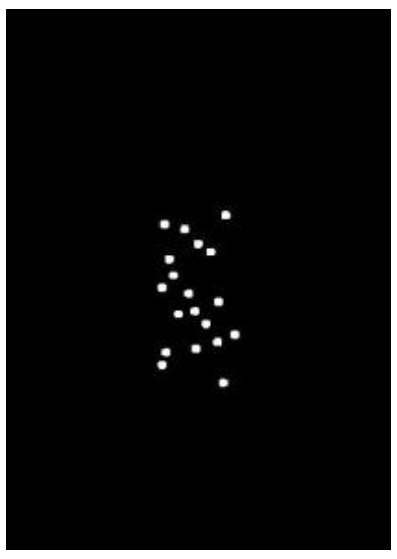

Fig.5 Final Processed Image

## V. Conclusion

Geometric camera calibration is one of the basic tasks in multi-camera systems in computer vision aspect. The character of the problem determines the requirements of the calibration method. In these type of systems, very high accuracy is needed. However, it is one of the most challenging task to achieve high accuracy because in getting high accuracy, measurements should be done very accurately.

In this paper a simple approach for multiple view reconstruction has been presented. However, according to the need of the problem more accurate & complex formulation can be done where each object in the scene will be calibrated for the reconstruction. But everything can be computed keeping our approach as the basis of computation.